\documentclass[conference]{IEEEtran}
\IEEEoverridecommandlockouts

\usepackage{cite}
\usepackage{amsmath,amssymb,amsfonts}
\usepackage{algorithmic}
\usepackage{graphicx}
\usepackage{textcomp}
\def\BibTeX{{\rm B\kern-.05em{\sc i\kern-.025em b}\kern-.08em
		T\kern-.1667em\lower.7ex\hbox{E}\kern-.125emX}}

\usepackage{multirow}

\usepackage{listings}
\usepackage{color}

\usepackage[table,xcdraw]{xcolor}
\usepackage[normalem]{ulem}
\useunder{\uline}{\ul}{}

\definecolor{dkgreen}{rgb}{0,0.6,0}
\definecolor{gray}{rgb}{0.5,0.5,0.5}
\definecolor{mauve}{rgb}{0.58,0,0.82}

\begin{document}

\title{A Reference Model for IoT Embodied Agents Controlled by Neural Networks}

\author{
    \IEEEauthorblockN{Nathalia Nascimento}
	\IEEEauthorblockA{\textit{David R. Cheriton School of Computer Science} \\
		\textit{University of Waterloo (UW)}\\
		Waterloo, Canada \\
		nmoraesdonascimento@uwaterloo.ca}
	\and
	\IEEEauthorblockN{Paulo Alencar}
	\IEEEauthorblockA{\textit{David R. Cheriton School of Computer Science} \\
		\textit{University of Waterloo (UW)}\\
		Waterloo, Canada \\
		palencar@uwaterloo.ca}
	\and 
	\IEEEauthorblockN{\hspace{2cm}Donald Cowan}
	\IEEEauthorblockA{\hspace{2cm}\textit{David R. Cheriton School of Computer Science} \\
	\hspace{2cm}	\textit{University of Waterloo (UW)}\\
	\hspace{2cm}	Waterloo, Canada \\
	\hspace{2cm}	dcowan@uwaterloo.ca}
	\and
	\IEEEauthorblockN{\hspace{2cm}Carlos Lucena}
	\IEEEauthorblockA{\hspace{2cm}\textit{Department of Informatics} \\
	\hspace{2cm}	\textit{Pontif{\'i}cia Universidade Catolica  (PUC-Rio)}\\
\hspace{2cm}		Rio de Janeiro, RJ, Brazil \\
	\hspace{2cm}	lucena@inf.puc-rio.br}
}

\maketitle

\begin{abstract}
Embodied agents is a term used to denote intelligent agents, which are a component of devices belonging to the Internet of Things (IoT) domain. Each agent is provided with sensors and actuators to interact with the environment, and with a `controller' that usually contains an artificial neural network (ANN). In previous publications, we introduced three software approaches to design, implement and test IoT embodied agents. In this paper, we propose a reference model based on statecharts that offers abstractions tailored to the development of IoT applications. The model represents embodied agents that are controlled by neural networks. Our model includes the ANN training process, represented as a reconfiguration step such as changing agent features or neural net connections. Our contributions include the identification of the main characteristics of IoT embodied agents, a reference model specification based on statecharts, and an illustrative application of the model to support autonomous street lights. The proposal aims to support the design and implementation of IoT applications by providing high-level design abstractions and models, thus enabling the designer to have a uniform approach to conceiving, designing and explaining such  applications.
\end{abstract}


\section{Introduction}\label{sec:introduction}

Approaches that incorporate agents, Machine Learning (ML) and Internet of Things (IoT) are being constantly developed \cite{savaglio2020agent}. These approaches have gained prominence because of the requirements of some IoT applications to support collective decision making in real-time within distributed smart environments \cite{savaglio2020agent}. In particular, in our previous research work (e.g. \cite{do2017fiot,nascimento2018iot}), we have proposed the use of embodied agents to denote intelligent agents which are embedded in the design of IoT applications. Each agent is provided with sensors and actuators to interact with the environment, and with a `controller' that usually contains an artificial neural network (ANN). Depending on the application task including an ANN, the agent is able to make decisions, predictions or classifications based on data collected from the environment.

However, there are still challenges related to the specification of intelligent embodied agent behavior in the face of an unpredictable and dynamic environment \cite{harel2020autonomics}. In order to address some of these challenges, there is a need for approaches that integrate Artificial Intelligence techniques, such as Machine Learning and Multiagent Systems, and Software Engineering (SE) approaches. According to Harel et al. (2019) \cite{harel2019labor}, SE techniques, such as statecharts \cite{harel1987statecharts}, can provide AI-based systems with intuitive and clear specifications that can result in systems that are easier to enhance and maintain compared to current uses of AI in system development.

In this paper, we introduce a reference model for IoT embodied agents and their interactions with the environment based on statecharts. Although there are many definitions for reference models, we assume a reference model is ``an abstract framework for understanding significant relationships among the entities of some environment" \cite{mackenzie2006reference}. Reference models can support the design and implementation of IoT applications by providing high-level design abstractions, thus making it easier to conceive and design these applications. In particular, we decomposed the part of the system based on neural networks into structural elements, which have been proposed by Harel et al. (2020) \cite{harel2020autonomics} as a solution towards making it easier to explain and verify ML-based systems.

The identification of the main characteristics of IoT embodied agents is based on three software approaches that we previously developed for IoT embodied agents: (i) a software framework for the development of embodied agents for IoT applications \cite{do2017fiot}; (ii) an approach to configure embodied agents based on the  environment \cite{nascimento2018iot}; and (iii) a method for testing embodied agents controlled by neural nets \cite{nascimento2019testing}.

The paper is structured as follows. Section \ref{sec:back} surveys the main concepts related to IoT embodied agents and their
requirements, taking previously published work into account. Section \ref{section:overview} provides high-level statechart models of embodied agents, and then we refine each component, describing, for example, the interaction between agent sensors and the input neurons of its ANN controller. Our model includes the ANN training process, representing it as configuration states such as selecting different sensors belonging to an agent or adjusting the weights of the neural network connections. Then, in Section \ref{section:example}, we extend these models to represent an illustrative application that uses IoT embodied agents. Section \ref{sec:conclusion} presents conclusions and future work.

\section{Background:} \label{sec:back}
This section provides an overview of IoT embodied agents and statecharts, which are the key concepts of our approach.

\subsection{IoT Embodied Agents}

Agents that can interact with other agents or the environment in which the applications are embedded are called {\itshape embodied agents} \cite{nascimento2018iot}. Examples of such agents can be found in areas such as autonomous robots and cyber-physical systems. The first step in creating an embodied agent is to design the agent body, which determines the agent interaction with application sensors and actuators and the corresponding signals that the agent will send and receive \cite{Nolfi2016}. As a second step, the agent is provided with a controller that instructs the agent how to behave based on signals from the agent. This controller is usually represented by an artificial neural network (ANN).

\begin{figure}[!htb]
	\centering
	\includegraphics[width=5.5cm]{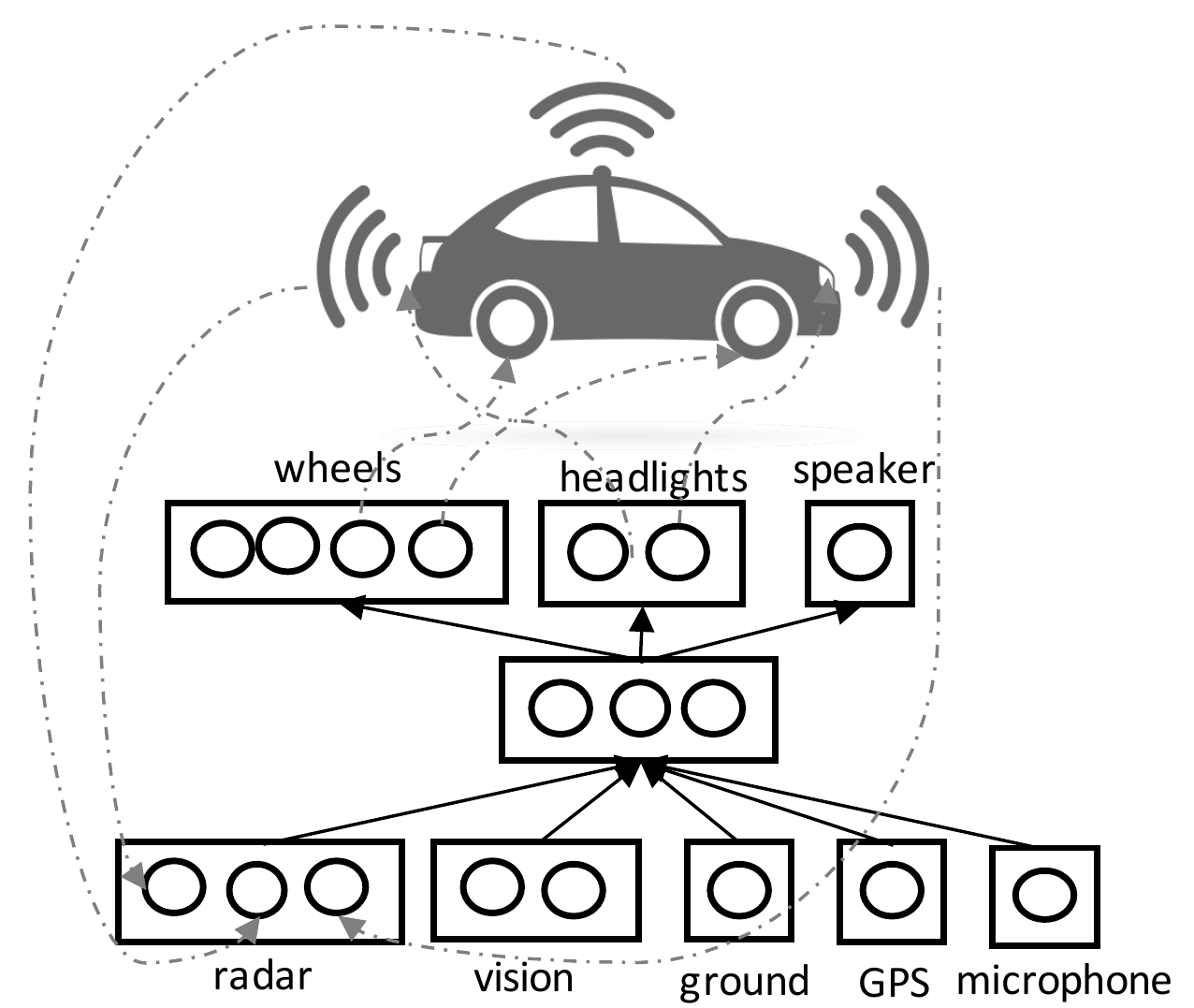}
	\caption{An example of an IoT embodied agent. }
	\label{figure:smartthing}
\end{figure}	

Figure \ref{figure:smartthing} illustrates an IoT embodied agent in an autonomous car scenario. In this example, the
body of the agent is a car with four wheels, GPS, headlights, and a speaker. As described previously, an embodied agent must have a local analysis architecture to sense the environment and behave accordingly. In this example, the autonomous car uses a neural net. There is an input neuron for each one of the car sensors and an output neuron for each one of the motors and actuators. The neuron output values may determine the direction of the wheels and whether the car turns on its headlights.

\subsection{Statecharts}

Statecharts is a formal method that extends the formalism of state machines and state diagrams with essentially three elements: hierarchy, concurrency and communication \cite{harel1987statecharts}. This formal method has been used to specify reactive systems, which are systems that respond to external and internal stimuli. The main elements of Statecharts are:

\begin{itemize}
    \item State and Events: a description of the dynamic behavior of a complex system \cite{harel1987statecharts};
    \item Transition: a transition is the connection between two states (source and destination) and can be represented by the triple: e[c] / a, in which ``e" is an event, ``c" is a condition, and ``a" is an action.
    \item Composition (or Clustering): introduces the XOR (exclusive-or) decomposition of states, which captures the property that, being in a state, the system must be in only one of its composite components;
    \item Orthogonality: independence and concurrency. It introduces the AND decomposition of states, which captures the property that, being in a state, the system must be in all of its AND components.
    \item History. In statecharts syntax, entering the history state means to enter the most recently visited state.
\end{itemize}

In short, state diagrams are simply directed graphs, with nodes denoting states, and arrows denoting transitions. Depending on the tool that is used to design statecharts, a state can be represented by different graphical notations. States are usually represented by rounded rectangles.

\section{A Reference Model for IoT Embodied Agents}\label{section:overview}

 In this section, we introduce high-level statechart models of embodied agents, and then we refine the boxes of each component such as the Body and Controller.

\subsection{Preliminary embodied agent concepts} \label{sec:initialconcepts}
Embodied agents are ``autonomous agents structurally coupled with their environment" \cite{franklin1997autonomous}. According to Quick et al. \cite{quick1999bots}, a system X is embodied in an environment E if perturbatory channels exist between the two. In other words, X is embodied in E if for  every  time  t  at which both X and E exist,  some subset of E's possible states have the capacity to perturb X's state, and some subset of X's possible states have the capacity to perturb E's state \cite{quick1999bots}. Accordingly, our proposed model must contain at least two components: embodied agents and the environment.

\subsubsection{Agent Body}
Basically, an agent body is composed of sensors and effectors. According to \cite{kinny2001reliable}, the agent ``interacts with its environment via interfaces of two types: sensors from which the agent receives events that carry information from its environment, and effectors by which the agent performs actions that are intended to affect its environment."

Auerbach and Bongard state that ``different parts of the robot body are responsible for different behaviors. For example, wheels or legs may allow for movement while a separate gripper allows for object manipulation" \cite{Auerbach:2009:EFS:1569901.1569915}. Based on the previous observations, our model describes the effects that the agent body has on its behavior.

\subsubsection{Agent Behavior} \label{sub:controlloop}
\begin{figure}[!htb]
	\centering
	\includegraphics[scale=0.42]{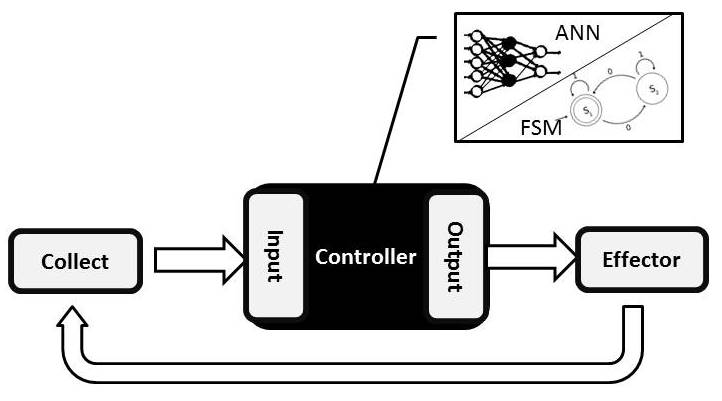}
	\centering
	\caption{Control loop executed by embodied agents.}
	\label{fig:controlloop}
\end{figure}

As we discuss in \cite{do2017fiot}, an autonomous embodied agent must execute a control loop with three key activities in sequence, namely: (i) collect  data; (ii) make decisions; and (iii) take actions. The task of data collection focuses on processing information coming from devices, such as reading data from input sensors. The collected data are used to set the inputs of the agent controller. Then, the controller processes a decision to be taken by the agent. This controller can be a finite state machine (FSM) or an ANN, as shown in Figure ~\ref{fig:controlloop}. Embodied agents act based on the controller output. An action such as an effector activity can be to interact with other agents, to send messages, or to set actuator devices, thus making changes to the environment.

\subsubsection{Agent Controller}
In addition, the intelligent behavior of an embodied agent arises out of the coupling of its body, its controller, and its environment. According to Auerbach and Bongard \cite{Auerbach:2009:EFS:1569901.1569915}, the complexity of an agent controller and body must match the complexity of a given task. However, more complex task environments require the agent to exhibit more different behaviors. Therefore, it is necessary to find the combination of agent body and controller that allows the agent to behave accordingly.

Based on the previous observations, our model also describes the effects of the agent controller on its behavior. In addition, it includes a new component to represent the task's  environment.

\subsection{Preliminary reconfiguration concept} \label{subsec:reconfig}
According to Karsai and Spanoudakis \cite{karsai1999model}, the reconfiguration mechanism must be associated with an evaluation module that represents how the system performance will be monitored and assessed, and how the evaluation result will affect the system architecture. In addition, the reconfiguration mechanism should be capable of interacting with the evaluator, being triggered by it.

The evaluation module measures a set of variables from the environment, and it can trigger the reconfiguration mechanism if it identifies a change in the set of measured variables.

Based on the previous observations, our model includes a new component to represent task evaluation, which is able to trigger the reconfiguration process.

\subsection{Statecharts Model}
Statecharts make it possible to view the description of the solution at different levels of detail, supporting a top-down behavioral specification. Therefore, it facilitates the understanding of a complex architecture. Other relevant characteristics are composition, hierarchy, and inter-component communications. Thus, our model can use transitions to explain the perturbations that occur between the environment and agents, which are the main components of embodied agents, as explained in Section \ref{sec:initialconcepts}.

\begin{figure}[!htb]
	\centering
	\includegraphics[width=8.6cm]{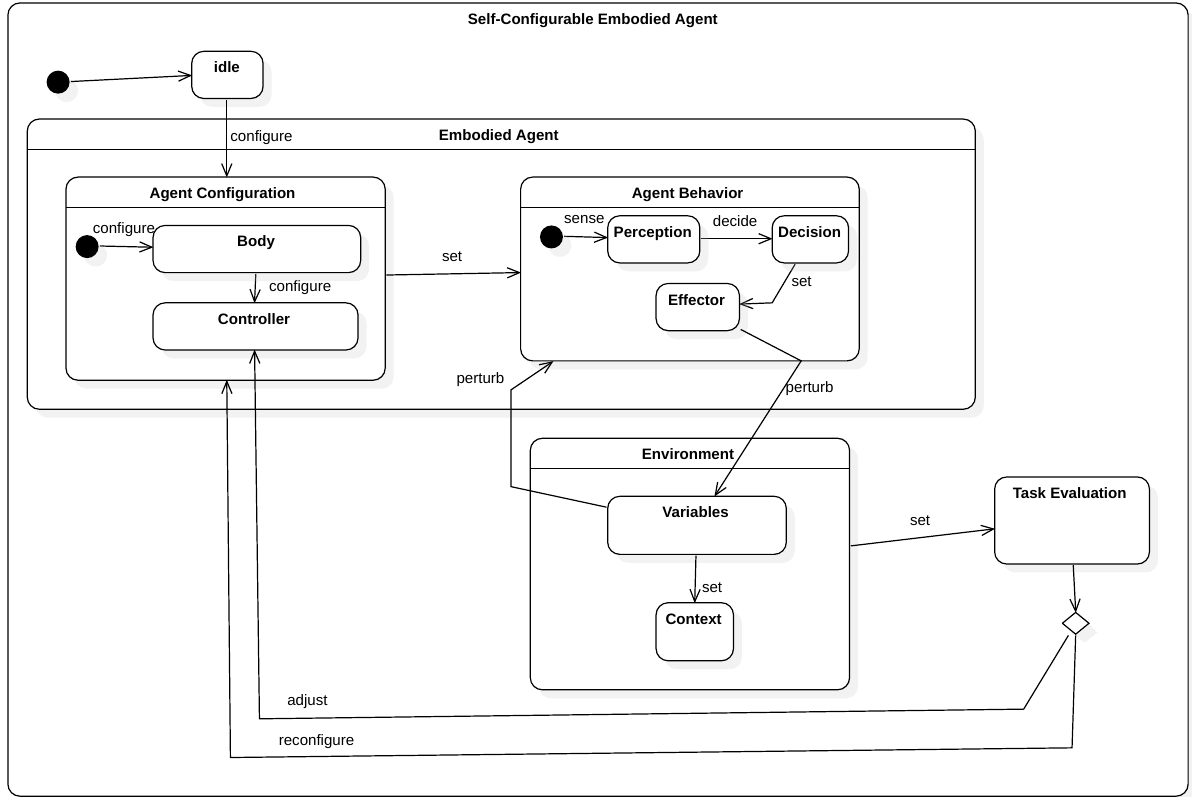}
	\caption{General statechart of IoT embodied agents - putting the main components together. }
	\label{figure:formal01}
\end{figure}

A high-level specification is given in Figure \ref{figure:formal01}, which contains the full statechart of an embodied agent. The main components of configurable embodied agents will be described in detail in the next subsections, in which we will look inside each one of these components.

The diagram in Figure \ref{figure:formal01} shows the states that are responsible for configuring an embodied agent based on its task. The agent comes out of the idle state when the system is initialized. As shown, we represented an embodied agent as a super state, and we assumed some physical and functional description of the system to describe its actions. Then, this state was hierarchically decomposed into agent configuration and agent behavior. The component of agent configuration is responsible for configuring the agent body and controller. Once the agent is configured, it is ready to behave in harmony with the environment. The agent behavior consists of three components: (i) perception; (ii) decision; and (iii) effect. As shown, the perception, decision and effector capabilities of the agent are directly related to how the body and controller of the agent are configured. To represent it, we use the transition ``set" between the components of configuration and behavior. Internal transitions have been omitted for simplicity.

Figure \ref{figure:formal01} also depicts inter-component communications between the agent behavior and environment components. As shown, the variables of the environment perturb the agent behavior. As Perception is the default state among Perception, Decision and Effector, the default way of entering this group of states is by the Perception state. In short, environment variables perturb the agent through perception. On the other hand, the effector state perturbs the environment. We will describe the details of these interactions in subsequent sections.

The unique way of entering the Task Evaluation component is through the environment component via ``set." As shown in Figure \ref{figure:formal01}, after evaluating the task, the system returns to the initial state, that is the agent configuration, resulting in a cycle. By entering the agent configuration state, the default option is to reconfigure the whole agent unless the transition ``adjust" is selected. If ``adjust" is selected, only the agent controller will be reconfigured.

\subsubsection{Agent Configuration - Body and Controller}

\begin{figure}[!htb]
	\centering
	\includegraphics[width=8.6cm]{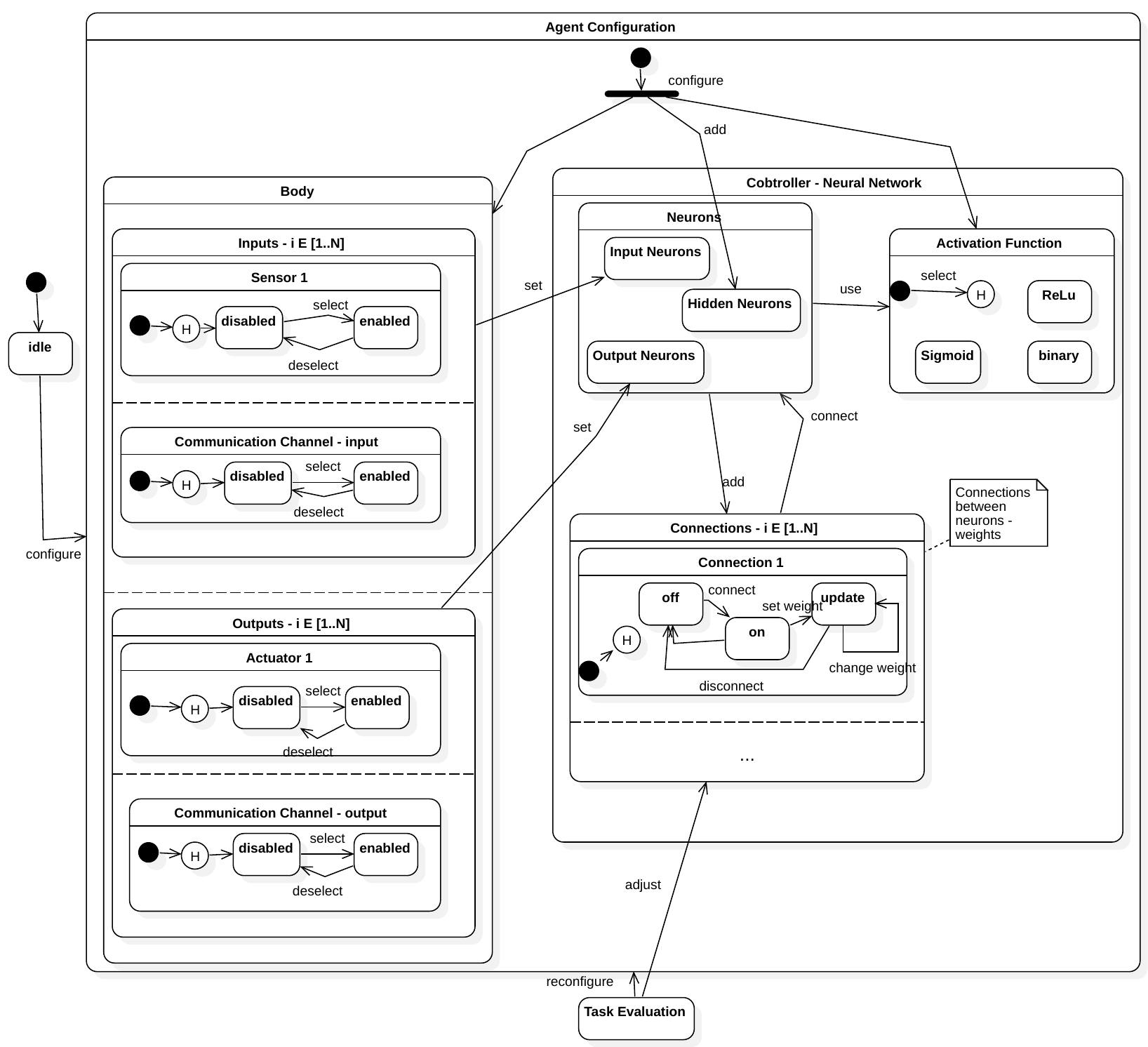}
	\caption{Body and Controller Configuration components of the embodied agent statechart. }
	\label{figure:formal02}
\end{figure}

A refinement of the agent configuration state yields Figure \ref{figure:formal02}. After examining the internals of the body component, we can see that the process of configuring the agent body consists of enabling or disabling some inputs, such as sensors, and outputs, such as actuators, which are represented with disabled/enabled substates. According to Harel (1987) \cite{harel1987statecharts}, ``an obvious application of orthogonality is in splitting a state in accordance with its physical subsystems." Thus, we represent the process of configuring agent body inputs and outputs as independent states. In other words, selecting the components to compose the body outputs does not depend on the components that are selected to compose its inputs. By default, all inputs and outputs are unselected, but we include a shallow history (``H") state in each component to allow the system to enter the most recently visited of the two, and enter ``disabled" if the system is there for the first time.

To configure the agent controller, there is a transition between the body inputs and the input neurons, and a transition between body outputs and output neurons. In other words, input neurons are set according to the enabled body inputs, and the output neurons are set according to the enabled body outputs. Each neuron may be connected to more than one neuron. So, a neuron adds a connection to the system and this connection is connected to a neuron (another or the same one). This statechart component represents the weight configuration of a neural network, in which the output of a neuron may enter another neuron as an input with a specific weight. After zooming-in to see more details of the agent controller, it is possible to observe that the ``adjust" transition segment, whose source is a state at the Task Evaluation component and directly associated with the connections component, is allowing the reconfiguration process to result only in enabling, disabling or updating connections.

\subsubsection{Agent Behavior} \label{sub:agentbehaviorstatechart}

\begin{figure}[!htb]
	\centering
	\includegraphics[width=8.6cm]{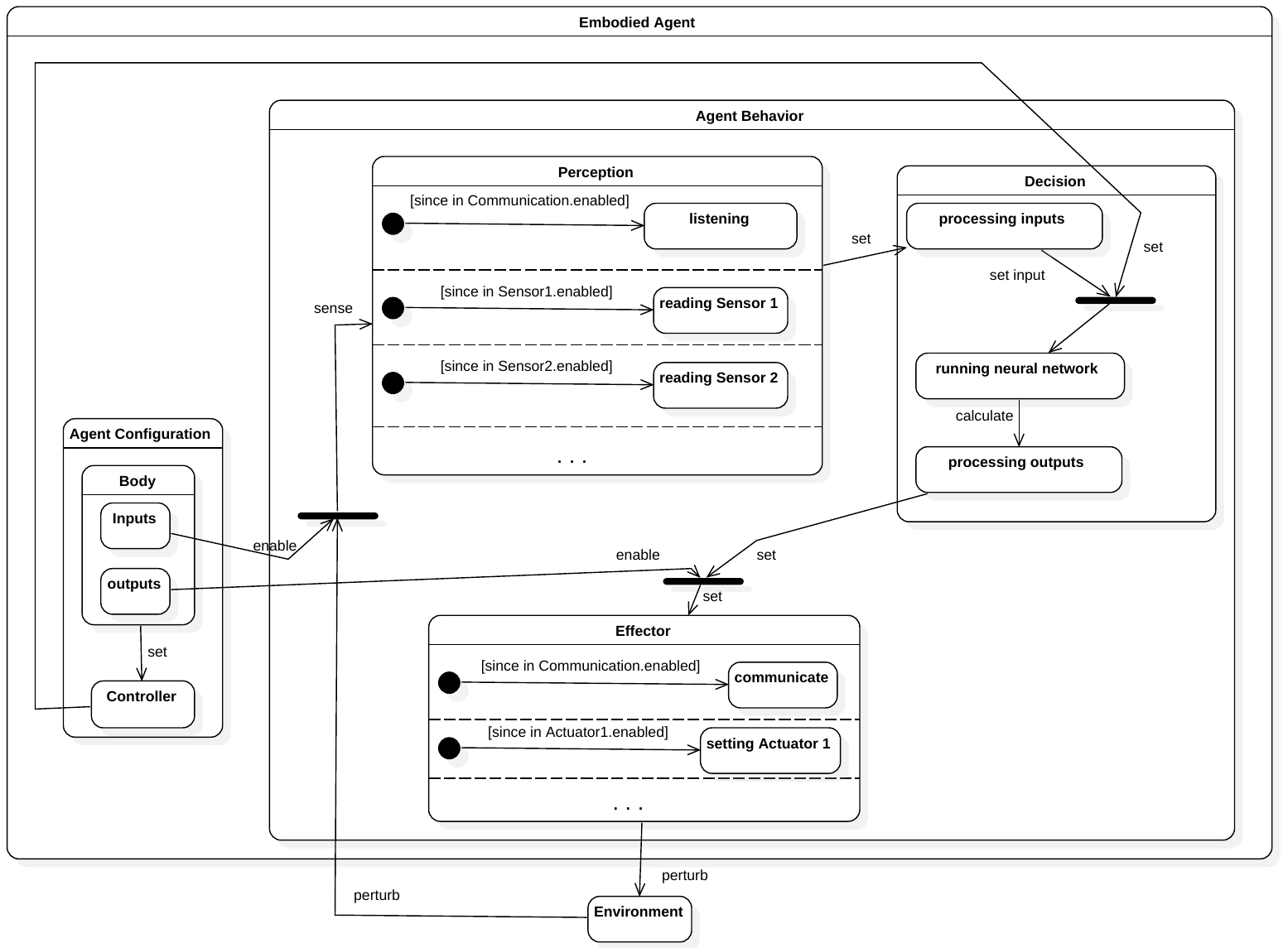}
	\caption{Behavior component of the embodied agent statechart.}
	\label{figure:formal03}
\end{figure}

As explained previously, the behavior of the agent varies based on the physical components that are operated by the agent and its controller. In short, the behavior of embodied agents is composed of three activities (see subsection \ref{sub:controlloop}): perception, decision and effect. According to Figure \ref{figure:formal03}, the perception task focuses on processing information coming from devices, such as reading data from input sensors. The collected data are used to set the inputs of the agent controller. Then, the controller processes a decision to be taken by the agent. Finally, the agent acts based on the controller output. An action (effector activity) can be to interact with other agents, to send messages, or to set actuator devices, thus making changes to the environment.

Selected inputs have an effect on the perception state, while outputs determine the effector state. For example, if the communication input is enabled, the agent will be able to listen to other agents. If the agent has a sensor A, the agent will be able to sense variable A of the environment; and if the communication output is activated, the agent will be able to communicate with other agents.

Note that we have used some joint states in Figure \ref{figure:formal03}. When a state emanates from a joint state, its source set consists of the sources of its constituent segments. For example, entering the substate ``running neural network"
depends on the incoming transitions from the states ``configuring controller" and ``processing inputs." In other words, how the collected inputs are processed to produce outputs depends on the neural network configuration. Consequently, the controller is crucial to the decision state. Perception and Effector substates also have more than one source. The transition segment ``perturb" from Environment and the segments ``enable" from Inputs are connected to enter the Perception substates, and this means that sensing a specific environment variable depends on the agent's sensors. Upon sensing some sensors on the ``Perception" component, the ``set" transition will be taken and the action ``processing inputs" will be carried out. Notwithstanding, entering Effector substates depends on the outgoing transitions from body outputs through actuators that are enabled and from the Decision state through output values that were calculated by the neural network. Results from Effector components perturb the environment, since some variables can be updated according to the agent's actions. For example, if the agent turns on the light, it will change the value of the brightness variable.

\subsubsection{Environment and Task Evaluation}

\begin{figure}[!htb]
	\centering
	\includegraphics[width=8.6cm]{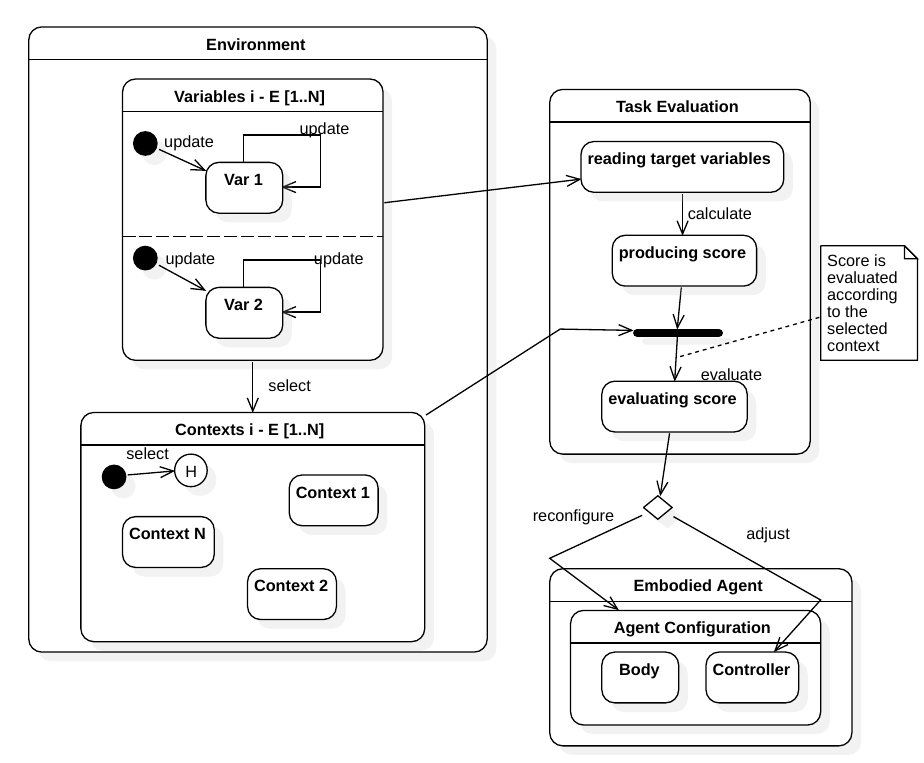}
	\caption{Statechart of Environment and Task Evaluation components. }
	\label{figure:formal04}
\end{figure}

The ``Environment" component is composed of variables and contexts \cite{nascimento2018context}, as depicted in Figure \ref{figure:formal04}. Variables represent perceivable characteristics of this environment. The ``Variables" component consists of N orthogonal substates, in which each substate is responsible for controlling a specific variable of the environment, such as brightness and humidity. They are also responsible for updating. Based on the variables' values, a specific context is selected. For example, supposing we have the contexts of ``day" and ``night", the current context will be selected according to the values of ``brightness" and ``time" variables.

In short, the ``Task Evaluation" component is responsible for examining the environment to investigate how the collection of embodied agents can be configured to deal with the system requirements and environmental changes. For this purpose, it will inspect specific variables of the environment in order to calculate a score. However, the significance of this score varies according to the current context, as we discuss in \cite{nascimento2018context}. In \cite{pezzulo2019making}, the authors illustrate a scenario with artificial agents where the evaluation policy varies according to the context. We represent this situation with a joint state. For example, if we are calculating energy consumption, it is expected that the energy that is spent during the night will be greater than during the day. So consuming 10kW during the day has a different impact than consuming this same energy value during the night. Further, the agent's reconfiguration will operate according to this evaluation. It can trigger the transition ``reconfigure" by restarting the whole process of configuring an agent, as shown in Figure \ref{figure:formal01}, or just trigger the transition ``adjust," reconfiguring only the neural network connections, as shown in Figure \ref{figure:formal02}.

\section{Illustrative Example: Autonomous Street Lights} \label{section:example}  
\begin{figure}[!htb]
	\centering
	\includegraphics[width=8.6cm]{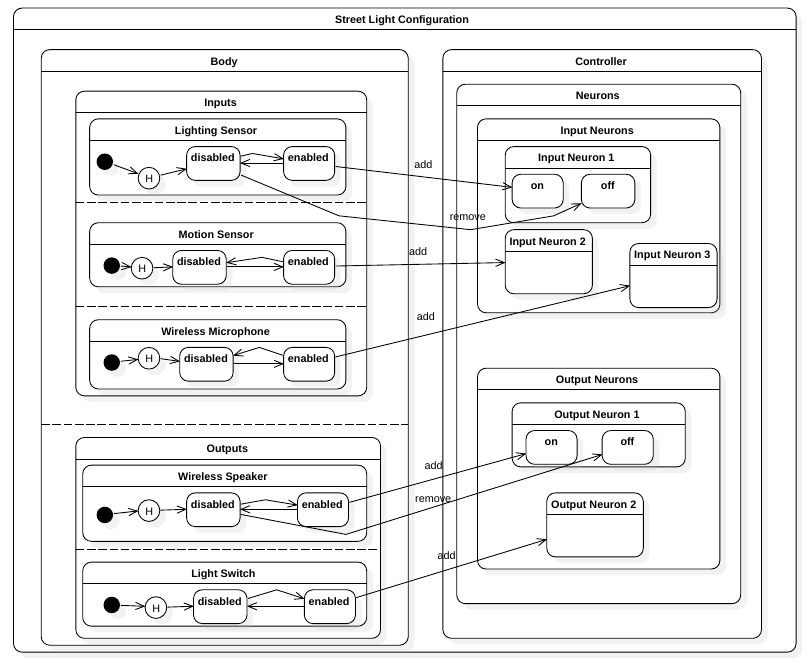}
	\caption{The Body and Controller components of the designed autonomous street lights.}
	\label{figure:formal05-1}
\end{figure}

This section introduces an illustrative example to show an application of the proposed reference model. In particular, we illustrate the statechart of embodied agents in a specific application scenario: autonomous street lights, in which each street light contains an embodied agent. In this scenario, each street light may contain a lighting sensor, motion sensor, wireless communication, wireless speaker and light switch apparatus. As we describe in Figure \ref{figure:formal05-1}, configuring the agent body consists of disabling or enabling some inputs and outputs. For example, we may have a street light agent containing only a lighting sensor as its input, and a light switch apparatus as its output; or we can create more robust street light agents by enabling them to communicate among themselves by means of wireless devices.

Configuring the agent controller also consists of disabling or enabling some components. As shown in Figure \ref{figure:formal05-1}, the number of neurons varies according to the sensors and actuators that were enabled. Accordingly, there will be an input neuron for each activated sensor and an output neuron for each activated output. To simplify this figure, we did not illustrate the Connections component shown in Figure \ref{figure:formal02}. Basically, if a neuron is in the ``on" state, that is it is active, it will be able to be connected to other neurons. For example, an output neuron can connect to an input neuron, consequently configuring a recurrent neural network.

\begin{figure}[!htb]
	\centering
	\includegraphics[width=8.6cm]{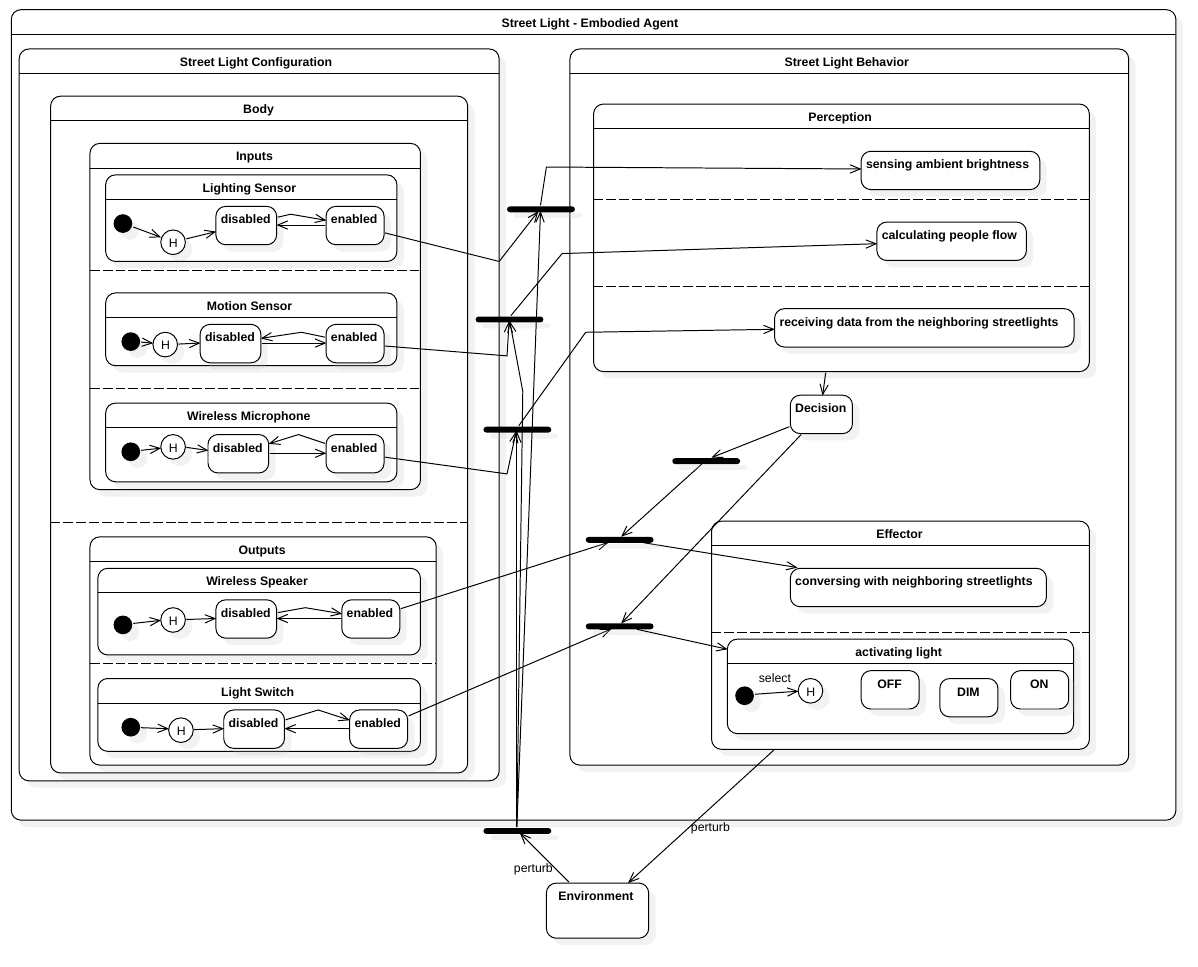}
	\caption{The Behavior component of the designed autonomous street lights.}
	\label{figure:formal05}
\end{figure}

As previously described, agent behavior is a consequence of the body and controller configuration, and the environment perturbation. If a specific sensor is enabled, the street light will be able to sense its specific variable, as described in Figure \ref{figure:formal03}. For example, as shown in Figure \ref{figure:formal05}, if the motion sensor is enabled, the agent will be able to calculate the flow of people in the environment. In the same way, if it has wireless communication, it may be able to receive communication signals from the other street lights.

Based on its outputs, the wireless speaker and the light switch apparatus, actions that can be taken by this agent are: ``conversing with neighboring streetlights" and ``activating light." For example, if an agent is able to activate the light, how many levels of brightness can this agent generate? As shown, to activate the light, the agent must select one of the following substates: OFF, DIM or ON. This selection depends on the results originated in the Decision state, as explained in subsection \ref{sub:agentbehaviorstatechart}.


\section{Conclusions and Future Work} \label{sec:conclusion}

We have identified the main characteristics of IoT embodied agents, and used them as a basis to propose a reference model based on statecharts. Meanwhile, we introduced a specification for embodied agents, hierarchically decomposing them into meaningful components. Our focus was not on providing a highly elaborate description, covering all possibly occurring concepts, but one that fits a broad variety of embodied agent models in different domains, as required by the wide range of IoT applications. Although our reference model proposal is at a high level of abstraction, we argue that it contributes particularly well to a view of IoT embodied agents, clarifying the relationship among their components, such as sensors and neural networks, and the complex and dynamic interactions between agents and their environments.

We have introduced this reference model so that it could serve to guide the development of software approaches that support the design, testing and implementation of embodied agents. Interesting future directions are to derive methods for explaining and verifying intelligent agent behavior and to extend the current approach to design IoT applications in specific domains such as smart cities.

\section*{Acknowledgment}
This work was supported by the Natural Sciences and Engineering Research Council of Canada (NSERC), and the Ontario Research Fund (ORF) of the Ontario Ministry of Research, Innovation, and Science, and the Centre for Community Mapping (COMAP).

\bibliographystyle{IEEEtran}
\bibliography{sigproc-2020,sigproc,quantifiedseke-bib,sigproc-adaptive}

\end{document}